\documentclass[10pt,twocolumn,letterpaper]{article}

\usepackage[pagenumbers]{main} 

\usepackage{graphicx}
\usepackage{amsmath}
\usepackage{amssymb}
\usepackage{booktabs}

\def\red#1{\textcolor{red}{#1}}
\def\blue#1{\textcolor{blue}{#1}}

\usepackage[pagebackref,breaklinks,colorlinks]{hyperref}

\usepackage[capitalize]{cleveref}
\crefname{section}{Sec.}{Secs.}
\Crefname{section}{Section}{Sections}
\Crefname{table}{Table}{Tables}
\crefname{table}{Tab.}{Tabs.}

\def\wacvPaperID{1382} 
\def\confName{WACV}
\def\confYear{2024}

\begin{document}

\title{Size-Variable Virtual Try-On with Physical Clothes Size}

\author{Yohei Yamashita ~~~ Chihiro Nakatani ~~~ Norimichi Ukita\\
Toyota Technological Institute\\
{\tt\small \{sd23501,ukita\}@toyota-ti.ac.jp}
}
\maketitle

\begin{abstract}
This paper addresses a new virtual try-on problem of fitting any size of clothes to a reference person in the image domain. While previous image-based virtual try-on methods can produce highly natural try-on images, these methods fit the clothes on the person without considering the relative relationship between the physical sizes of the clothes and the person. Different from these methods, our method achieves size-variable virtual try-on in which the image size of the try-on clothes is changed depending on this relative relationship of the physical sizes. To relieve the difficulty in maintaining the physical size of the closes while synthesizing the high-fidelity image of the whole clothes, our proposed method focuses on the residual between the silhouettes of the clothes in the reference and try-on images. We also develop a size-variable virtual try-on dataset consisting of 1,524 images provided by 26 subjects. Furthermore, we propose an evaluation metric for size-variable virtual-try-on. Quantitative and qualitative experimental results show that our method can achieve size-variable virtual try-on better than general virtual try-on methods.
\end{abstract}

\section{Introduction}
\label{sec:intro}

\begin{figure}[t]
  \begin{center}
  \includegraphics[width=\columnwidth]{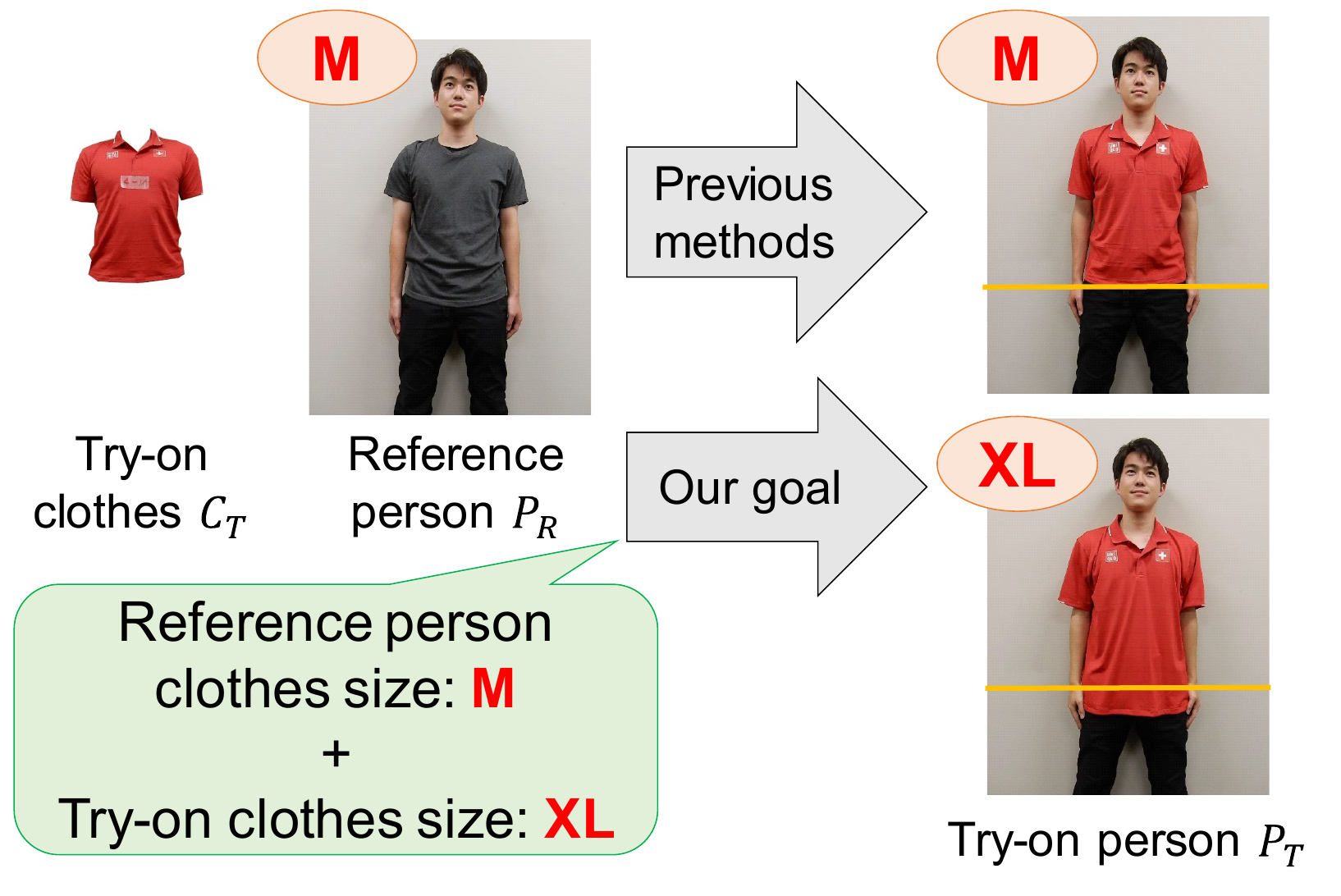}
  \end{center}
  \caption{
  Comparison between previous methods and our method. (i) Previous methods~\cite{viton,acgpn} only generate the try-on image in which the clothes size is the same as the one of the reference person. (ii) A user can change the try-on cloth size in our method.
  }
  \label{fig:top}
\end{figure}

The apparel industry is rapidly shifting to e-commerce, and its market size is expanding.
However, customers cannot try on clothes online.
The alternative is virtual try-on~\cite{old1,old2,old3,old4}.
We can try on any clothes in images.
The quality of virtual try-on is improved by the success of image synthesis tasks such as image harmonization~\cite{harmonize1,harmonize2,harmonize3}, image inpainting~\cite{edgeconnect,gated,inpainting}, and image editing~\cite{scfegan,maskgan,edo}.

In the 3D virtual try-on task~\cite{pixel2surf,3d,M3D-VTON}, a 3D avatar~\cite{smpl,mgn} generated from a user tries on any clothes by rendering the texture of the clothes on the avatar. 
However, generating a realistic 3D avatar for each user takes much time and cost.

In contrast to the 3D virtual try-on task, the 2D virtual try-on task only requires the images of the user and clothes. 
This image-based virtual try-on task can be categorized into  (i) warping-based virtual try-on and (ii) Image Style Transfer-based (IST-based) virtual try-on.

Warping-based methods~\cite{viton,acgpn} consist of two steps. 
First, the silhouette of the clothes is estimated as a mask in the images of the try-on clothes and the reference person (i.e., a user).
The mask of the try-on clothes is warped to fit with that of the reference person's clothes
without considering the physical size of the try-on clothes (e.g., small, medium, and large), as shown in the upper row of Fig.~\ref{fig:top}.

While the warping-based methods cannot control the size of the try-on clothes, IST-based methods~\cite{tryongan} potentially allow us to control the clothes size by adjusting the weights of conditioning for image style transfer.
However, it is not easy to appropriately control the clothes size (so that the try-on clothes fits with its physical size in the image) because the relationship between the conditioning weights and the physical size is unknown and highly complex.

In summary, the previous warping- and IST-based 
virtual try-on
methods cannot reflect the physical size of try-on clothes in the try-on image.
To address this issue, we propose size-variable virtual try-on with the physical size, as shown in the lower row of Fig.~\ref{fig:top}.
Our contributions are as follows:
\begin{enumerate}
\item {\bf Size-variable virtual try-on:}
This paper defines a new problem, namely size-variable virtual try-on.
Its goal is to fit the try-on clothes with the human body image by taking into account their physical sizes given as a user's preference.

\item {\bf Size-variable mask deformation network:} 
Size-variable virtual try-on is divided into two sub-tasks, size-variable mask deformation and texture rendering within the deformed mask.
This paper focuses on the former (MDN: Mask Deformation Network in Fig.~\ref{fig:warping}), while the latter is done with existing methods (TPS: Thin Plate Spline and CFN: Content Fusion Network in Fig.~\ref{fig:warping}).
Our MDN maintains the whole silhouette of the try-on clothes while adjusting its image size in accordance with its physical size given by a user.
Our method achieves this silhouette maintenance and size adjustment by focusing on the residual between the cloth silhouettes in the reference and try-on images.
%

\item {\bf Size evaluation:} 
Size-variable virtual try-on is a new problem, so we propose a new evaluation metric, namely the Size Evaluation Metric (SEM).
SEM evaluates the size differences of hem and sleeve areas that are important for size-variable virtual try-on.

\item {\bf Size-variable virtual try-on dataset:}
We also develop a new dataset for size-variable virtual try-on.
\end{enumerate}

\section{Related Work}
\label{sec:related_work}

\subsection{Warping-based Virtual Try-On}
\label{subsec:warped_vton}

\begin{figure}[t]
  \begin{center}
    \includegraphics[width=\columnwidth]{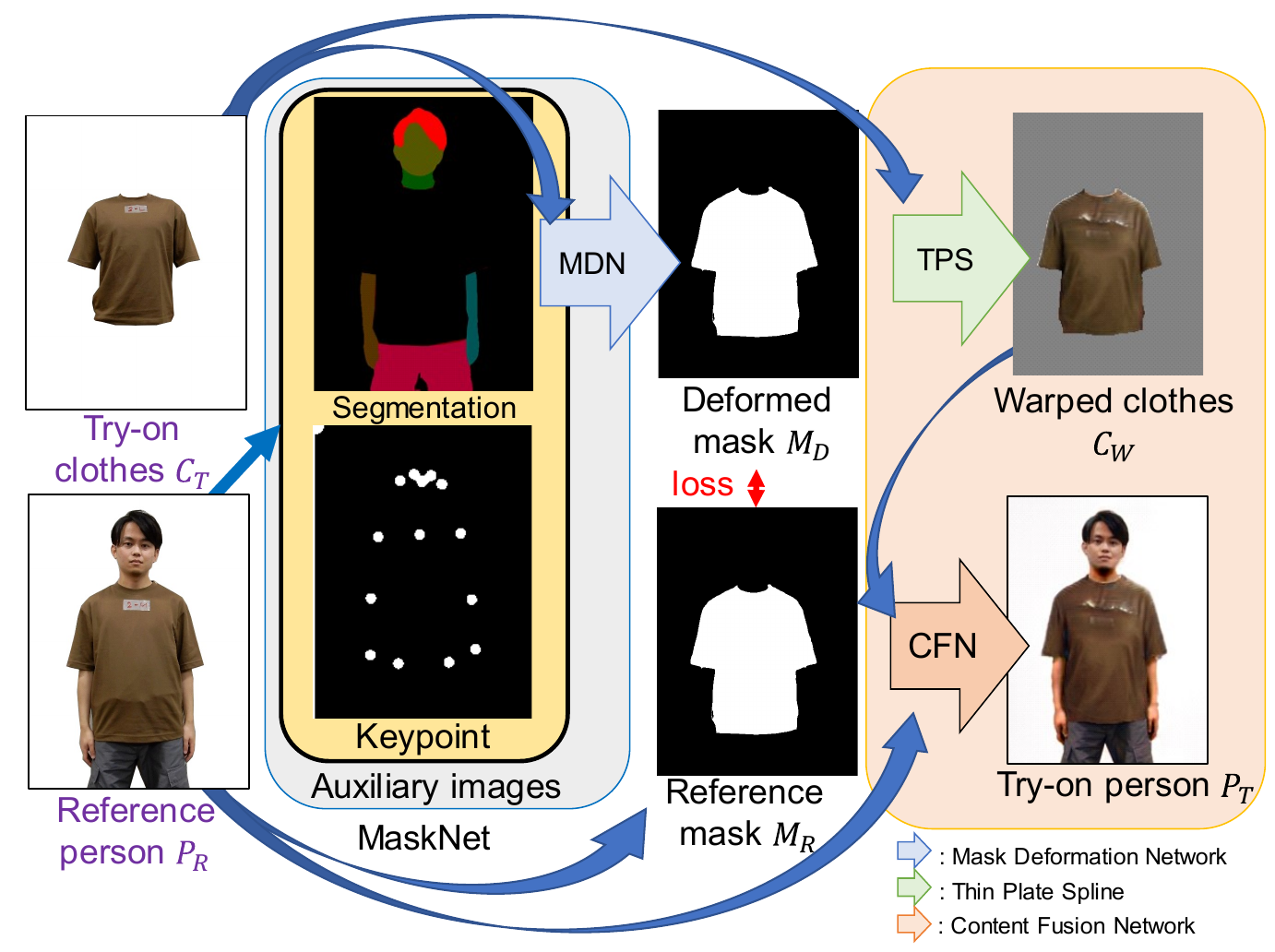}
  \end{center}
  \caption{Overview of warping-based virtual try-on methods. 
  The try-on mask is estimated from the auxiliary and clothes images. 
  The clothes are warped to fit the mask and integrated with the reference person image to generate the try-on image.
  }
  \label{fig:warping}
\end{figure}

Figure~\ref{fig:warping} shows the two-stage pipeline of general warping-based methods~\cite{viton,cp-viton,vtnfp}. 
(i) The mask of the try-on clothes ($C_{T}$) is deformed to fit with the reference person image ($P_{R}$).
To make this deformation easier, the segmentation image and the person key-points are extracted as the auxiliary images from $P_{R}$ and fed into MDN with $C_{T}$.
(ii) $C_{T}$ is warped to fit with the deformed mask ($M_{D}$) by TPS~\cite{tps}, and then the warped try-on clothes ($C_{W}$) and $P_{R}$ are fused to produce the try-on person image ($P_{T}$) by CFN.

However, TPS sometimes causes a large erroneous deformation on $C_{W}$.
In~\cite{acgpn,dicton}, this problem is relieved by segmenting $P_{T}$ to the generated and original pixels so that the pixel values in the original pixels are copied from $P_{R}$.
While these methods~\cite{acgpn,dicton} can preserve the quality of the original pixels, the quality of the generated pixels in $P_{T}$ is degraded if these methods are applied to high-resolution images.
VITON-HD~\cite{viton-hd} iteratively updates the segmentation image and increases the resolution of $P_{T}$ for high-resolution virtual try-on. 
While such segmentation-based methods can be affected by erroneous segments, knowledge distillation-based methods~\cite{wuton,parserfree,style-af} can generate $P_{T}$ without segmentation.
%

In all of these methods, the physical size of the try-on clothes (e.g., hem and sleeve) is not explicitly addressed. 
Unlike these methods, our method estimates a size-variable try-on mask according to the physical size of the clothes.

\subsection{Image Style Transfer-based Virtual Try-On}
\label{subsec:style_vton}

Image Style Transfer (IST) such as StyleGAN~\cite{stylegan,stylegan2,image2stylegan,image2stylegan++} can generate images from the learned disentangled latent space. 
The disentangled latent space allows us to edit the specific regions of the generated image (e.g., hair length and eye color). 
StyleGAN is extended to virtual try-on in TryOnGAN~\cite{tryongan}.
TryOnGAN generates $P_{T}$ by fusing disentangled style vectors representing the attributes of a person in an image and clothes in another image. 
However, TryOnGAN optimizes the style vectors of the clothes so that the try-on clothes in $C_{T}$ fit with the body shape in $P_{R}$ without taking into account the physical size of the clothes.

IST-based methods such as TryOnGAN can be extended to change the size of try-on clothes in $P_{T}$ by changing noise given to the disentangled style vectors.
However, the relationship between the image and physical sizes of try-on clothes is unknown.
Different from such IST-based methods, our method estimates the try-on mask from the physical size of try-on clothes for size-variable virtual try-on.

\begin{figure}[t]
  \begin{center}
  \includegraphics[width=\columnwidth]{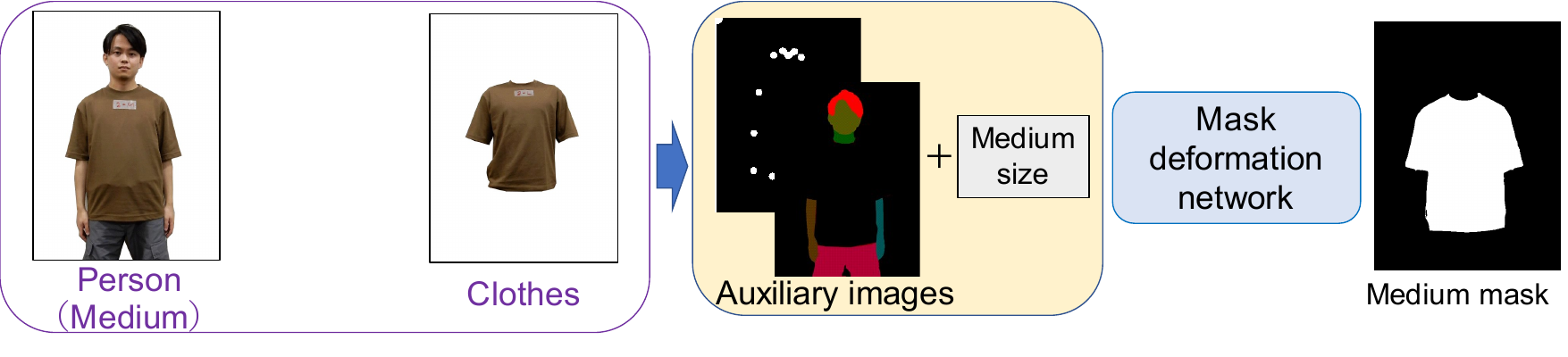}
  (a) Straightforward extension of previous methods
  \end{center}
  \begin{center}
  \includegraphics[width=\columnwidth]{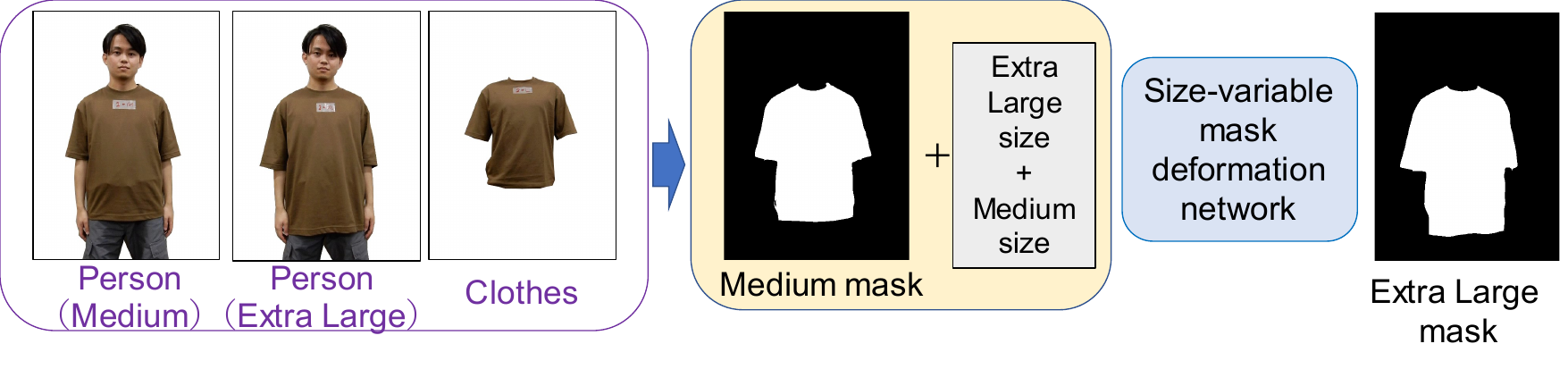}
  (b) Our method
  \end{center}
  \caption{Two different approaches of mask deformation.
  (a) Extension of previous methods.
  The size of the try-on clothes is used for training MDN with auxiliary images in previous methods.
  (b) Our method.
  The paired person images in which each person wears different sizes of the same clothes are used in training.
  }
  \label{fig:sv_vto_comp}
\end{figure}


\section{Size-Variable Virtual Try-On}
\label{sec:size_vton}

For size-variable virtual try-on, we collected a new dataset introduced in Sec.~\ref{subsec:prop_dataset}. 
Our proposed size-variable MDN is described in Sec.~\ref{subsec:prop_mask deformation network}. 
Furthermore, we propose a new evaluation metric for size-variable virtual try-on (Sec.~\ref{subsec:prop_eval}).

For size-variable virtual try-on, a straightforward scheme is to provide the physical size of try-on clothes as auxiliary cues to a previous virtual try-on method, as shown in Fig.~\ref{fig:sv_vto_comp} (a).
However, such a straightforward scheme is not effective because only the size of clothes (as numerical parameters, which are indicated by ``Parameters (Try-on size)'' in Fig.~\ref{fig:sv_vto_comp} (a)) is not enough informative to generate the size-aware clothes mask.
On the other hand, our method generates the size-aware clothes mask by adding the clothes image of the target size (which is indicated by ``Person (Try-on size)'' in Fig.~\ref{fig:sv_vto_comp} (b)) as well as the sizes of reference clothes and try-on clothes (as numerical parameters, which are denoted by ``Parameters (Reference size)'' and ``Parameters (Try-on size),'' respectively) in training.
This training is achieved by conditioning our MDN by the preferred size of the clothes so that the MDN output coincides with the mask of the clothes of the preferred size.

\begin{figure}[t]
  \begin{center}
     \includegraphics[width=\columnwidth]{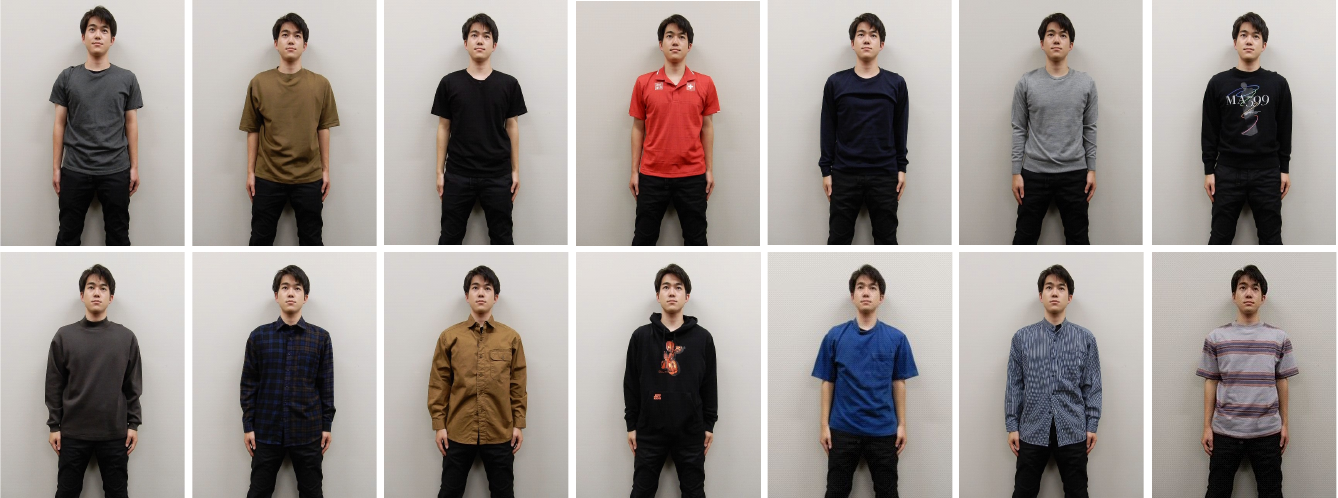}
     \caption{Sample images of 14 clothes.
     }
     \label{fig:dataset_samples}
  \end{center}
\end{figure}

\begin{figure}[t]
  \begin{center}
  \includegraphics[width=\columnwidth]{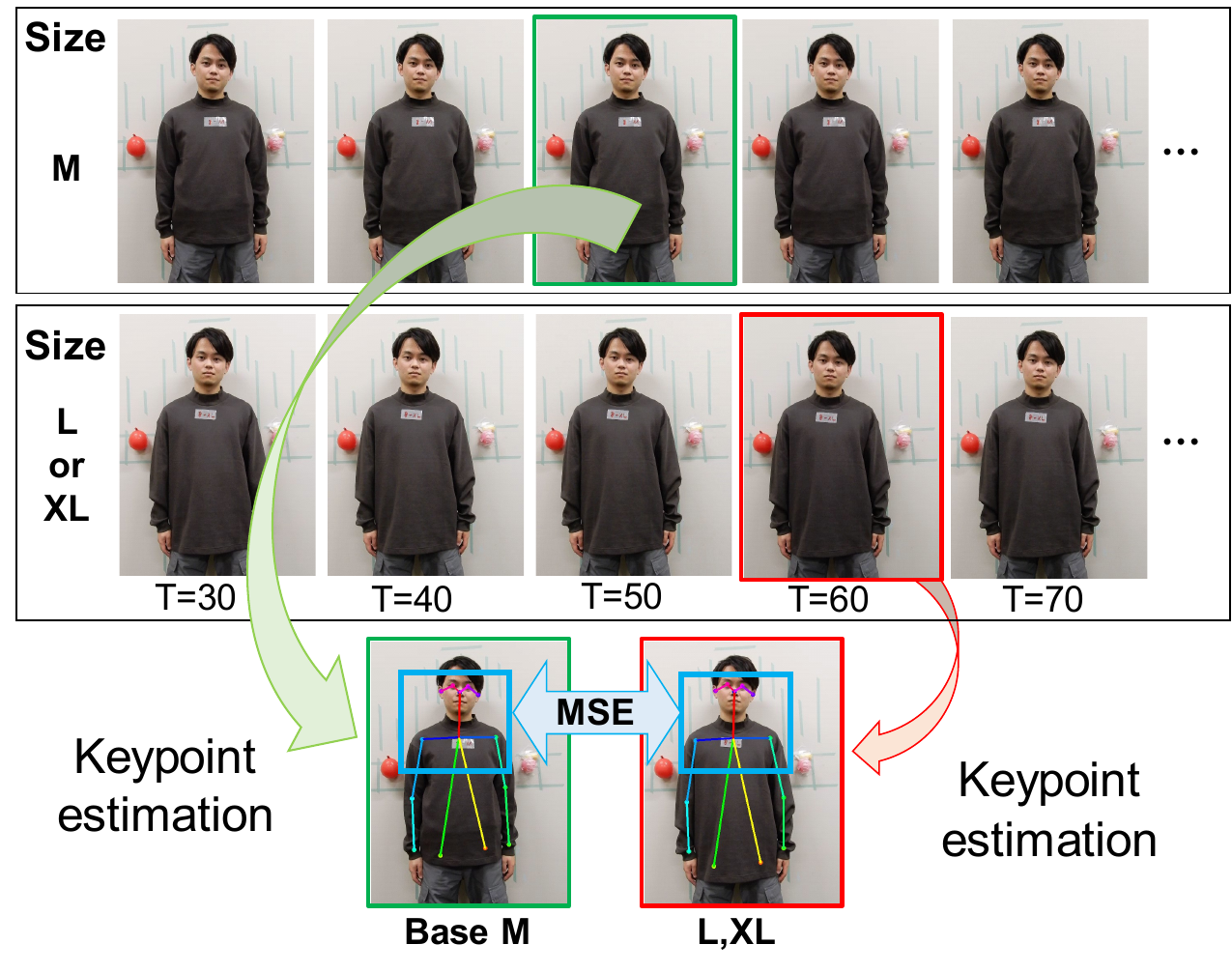}
  \end{center}
  \caption{Posture matching for collecting image pairs, in each of which each subject's postures are similar.
  This matching is done with the Mean Squared Error (MSE) computed between the sets of several body key-points.
  }
  \label{fig:dataset_pos_match}
\end{figure}

\begin{figure*}[t]
  \begin{center}
     \includegraphics[width=\textwidth]{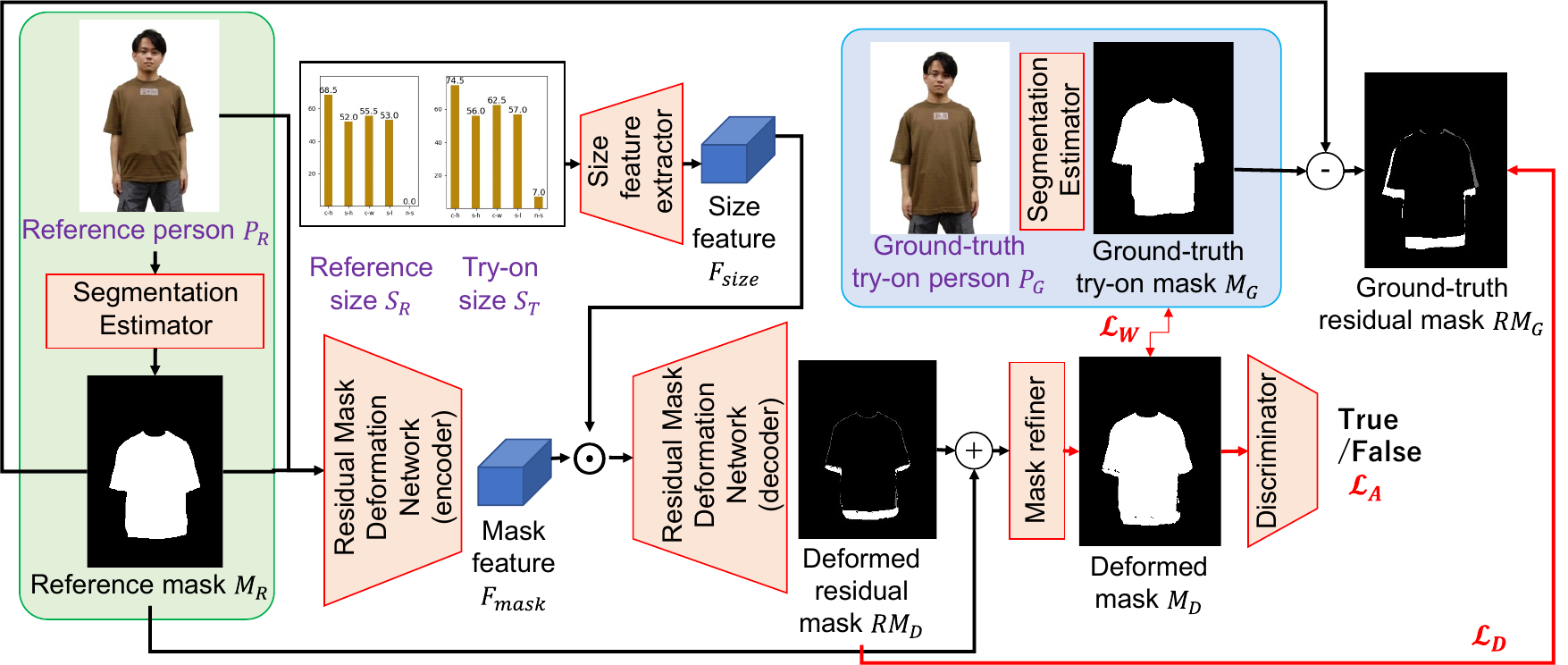}
     \caption{Overview of our size-variable mask deformation network. 
     The final output mask ($M_{D}$) is estimated from $P_{R}$ and the sizes of the reference and try-on clothes ($S_{R}$ and $S_{T}$, respectively).
     To focus on the small difference between the reference mask $M_{R}$ and try-on mask $M_{G}$, the residual between these two masks is estimated as the intermediate output ($RM_{D}$).
     $RM_{D}$ is fed into the Mask Refiner (MR). 
     }
     \label{fig:mask deformation network}
  \end{center}
\end{figure*}

\subsection{Size-variable Virtual Try-on Dataset}
\label{subsec:prop_dataset}

\noindent{\bf Motivation.}
While the Zalando Dataset~\cite{viton} has been widely used for the virtual try-on task, this dataset only contains the pairs of clothes and person images without the clothes size.
This disadvantage motivates us to collect a new dataset that contains the clothes size that can be used for size-variable virtual try-on.

\noindent{\bf Overview.}
In our dataset, each data is a pair of images in which observed clothes are the same except for physical size (e.g., ``Person (Reference size)'' and ``Person (Try-on size)'' in Fig.~\ref{fig:sv_vto_comp} (b)).
Each image is annotated with the physical clothes size.
The size parameters of each clothes are ``Body length back,'' ``Sleeve length,'' ``Shoulder width,'' ``Body width,'' and ``Neck size.''
A set of these size parameters is represented as a 5D vector.
Our dataset is generated from 1,524 images of 26 subjects with 14 types of clothes shown in Fig.~\ref{fig:dataset_samples}.
From the 1,524 images, 3,746 image pairs are collected so that the person's postures are almost the same in each image pair.
All the paired data are split into 3,121 training, 529 validation, and 96 test data.
In the 96 test data, there are 24 new clothes data, in which subjects wear new try-on clothes that are not included in the training data, and 72 new person data, in which subjects are excluded from those in the training data.

The distance from a camera to a subject was almost fixed in our dataset images.
In real application scenarios, on the other hand, this distance may differ depending on the image capturing condition.
This gap can be suppressed by rescaling/normalizing an input image in inference according to the ratio between the pixel and physical sizes (i.e., heights) of a user observed in the input image because the ratio in the dataset image is known.

\noindent{\bf Dataset collection.}
While it is required to spatially align the human postures between the two images in each image pair for our proposed method, this human posture alignment is not easy because it is difficult for a subject to be in the same posture in  two shots.
Therefore, we propose posture matching to collect the paired images in which the postures of a person are almost the same, as shown in Fig.~\ref{fig:dataset_pos_match}.
This posture matching consists of the following four steps. 
(i) Videos of each subject wearing different-sized same-type clothes are captured.
The subject is requested to be in the same posture between the videos of the different-sized clothes.
In each video pair (e.g., upper and lower videos in Fig.~\ref{fig:dataset_pos_match}), the following steps (ii), (iii), and (iv) are done.
(ii) The body key-points of the subject are estimated in all frames in the pair videos.
In each of all possible frame pairs between the pair videos, the posture similarity between the sets of the key-points above the shoulders (which are within the blue rectangles in Fig.~\ref{fig:dataset_pos_match}) is evaluated.
This is because the key-points below the shoulders tend to  differ even if the subject tries to be in the same posture.
This posture similarity is measured as the Mean Squared Error (MSE) between the set of the key-points.
(iii) The frames in which the MSE is smallest are selected as a matched image pair, as enclosed by the green and red rectangles in Fig.~\ref{fig:dataset_pos_match}.
We call the dataset collected by the above protocol ``BaseDataset.''
(iv) For further reducing the spatial displacement between the pair images in the BaseDataset, one of the pair images is warped in order to spatially align the sets of the key-points between the two images by projective transformation.
This dataset is called ``ProjDataset.''


\subsection{Size-variable Mask Deformation Network}
\label{subsec:prop_mask deformation network}

The detail of our proposed size-variable mask deformation network, which is ``MDN'' in Fig.~\ref{fig:warping}, is shown in Fig.~\ref{fig:mask deformation network}.
For training this network, the pair images ($P_{R}$ and $P_{G}$) annotated with the sizes of clothes in these images ($S_{R}$ and $S_{T}$) are given, as described in Sec.~\ref{subsec:prop_dataset}. 
These inputs are fed into the network to estimate the deformed mask $M_{D}$ used for warping-based virtual try-on, shown in Fig.~\ref{fig:warping}.

\noindent{\bf Architecture.}
The image of reference person $P_{R} \in\mathbb{R}^{3 \times H \times W}$ is fed into a segmentation estimator (SE) to obtain its reference mask $M_{R} \in\mathbb{R}^{1 \times H \times W}$.
$H$ and $W$ denote the height and width of the image, respectively.
$P_{R}$ and $M_{R}$ are fed into the Residual Mask Deformation Network (RMDN) to extract their feature map $F_{mask} \in\mathbb{R}^{C \times H' \times W'}$ where $C$, $H'$, and $W'$ denote the dimension of $F_{mask}$.
The size parameters of clothes in $P_{R}$ and $P_{G}$ (i.e., $S_{R} \in\mathbb{R}^{S}$ and $S_{T} \in\mathbb{R}^{S}$ where $S$ denotes the number of parameters representing the size of clothes) are fed into the Size Feature Extractor (SFE) to extract their feature map ($F_{size} \in\mathbb{R}^{C \times H' \times W'}$).
Then, $F_{mask}$ and $F_{size}$ are fused by elementwise multiplication. 
The fused feature map is decoded to estimate the residual mask $RM_{D} \in\mathbb{R}^{1 \times H \times W}$. 
$RM_{D}$ and $M_{R}$ are elementwise added to obtain the whole mask. 
Finally, the whole mask is fed into the Mask Refiner (MR) consisting of two convolutional layers to obtain the deformed mask $M_{D} \in\mathbb{R}^{1 \times H \times W}$.
Our RMDN allows us to focus on the minor difference between $M_{R}$ and $M_{G}$.

\noindent{\bf Ground-truth Mask Generation.}
The ground-truth try-on mask $M_{G} \in\mathbb{R}^{1 \times H \times W}$ and the ground-truth residual mask $RM_{G} \in\mathbb{R}^{1 \times H \times W}$ are generated from the ground-truth try-on person image $P_{G}$. 
As with $M_{R}$, $M_{G}$ is generated by SE.
$RM_{G}$ is generated from the elementwise subtraction between $M_{R}$ and $M_{G}$.
Both $M_{G}$ and $RM_{G}$ are used to train the whole network by the following loss functions.

\noindent{\bf Loss functions.}
The whole network is trained on the following loss functions:
\begin{equation}
\mathcal{L}= \lambda_{W}\mathcal{L}_{W} + \lambda_{D}\mathcal{L}_{D} + \lambda_{A}\mathcal{L}_{A},
\label{eq:loss}
\end{equation}
where $\mathcal{L}_{W}$, $\mathcal{L}_{D}$, and $\mathcal{L}_{A}$ denote the Weighted Binary Cross Entropy loss, the Dice loss~\cite{dice}, and the Adversarial loss~\cite{pix2pix}, respectively.
$\lambda_{W}$, $\lambda_{D}$, and $\lambda_{A}$ are their weights.
$\mathcal{L}_{W}$ is computed with $M_{D}$ and $M_{G}$.
$\mathcal{L}_{D}$ is computed with $RM_{D}$ and $RM_{G}$ to focus on the minor difference between $M_{R}$ and $M_{G}$.
$\mathcal{L}_{A}$ is computed with $M_{D}$ and $M_{G}$ so that $M_{D}$ as a fake data gets close to $M_{G}$ as a true data. $\mathcal{L}_{A}$ can improve the reality of the boundary of $M_{D}$.

\noindent{\bf Inference.}
In inference, $P_{R}$, $S_{R}$, and $S_{T}$ are given and fed into our size-variable mask deformation network.
Its output $M_{D}$ is used to generate a warped clothes image 
$C_{W}$ and a try-on image $P_{T}$ by a general warping-based virtual try-on method, as described in Sec.~\ref{subsec:warped_vton} and shown in Fig.~\ref{fig:warping}.

\begin{figure}[t]
  \begin{center}
    \includegraphics[width=\columnwidth]{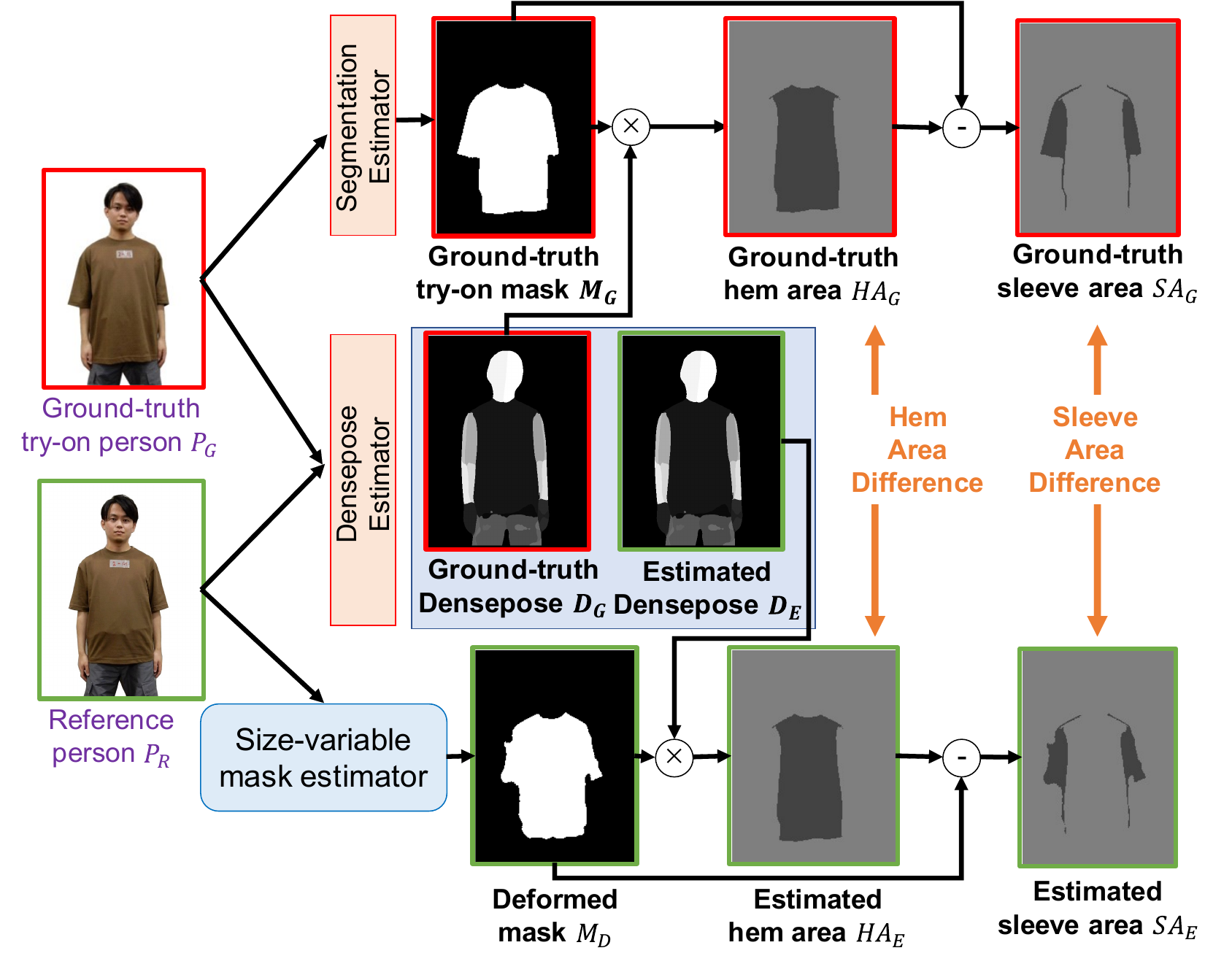}
  \end{center}
  \caption{
  Size Evaluation Metric (SEM).
  }
  \label{fig:eval}
\end{figure}


\subsection{Size Evaluation Metric}
\label{subsec:prop_eval}

While each of $M_{D}$ and $M_{G}$ is estimated as a heatmap image, the pixelwise difference between these two masks cannot be directly used for evaluating how much $M_{D}$ looks like $M_{G}$ because $M_{D}$ and $M_{G}$ are misaligned, as mentioned in Sec.~\ref{subsec:prop_dataset}.
Furthermore, evaluation using all pixels cannot focus on the changes of areas important for size-variable virtual try-on (e.g., sleeves and hems).
To resolve these problems, we propose the Size Evaluation Metric (SEM), as shown in Fig.~\ref{fig:eval}.

For SEM, the torso hem and sleeves are considered to be areas important for size-variable virtual try-on.
These areas are identified based on human body-part segments.
These segments are detected by Densepose~\cite{densepose}.
$D_{R}$ and $D_{G}$ denote the Densepose heatmaps estimated from $P_{R}$ and $P_{G}$, respectively.
In the ground-truth image, its torso area ($T_{G}$) is identified to be the pixelwise multiplication between $M_{G}$ and the sum of the torso and upper-legs segments in $D_{G}$.
In the same manner, the torso area in the reference image ($T_{D}$) is identified with $M_{D}$ and $D_{R}$.
The sleeve areas are identified to be the pixelwise subtraction between the clothes mask and the torso area.
The sleeve areas are denoted by $S_{G}$ and $S_{D}$.
The differences between ``$T_{G}$ and $T_{D}$'' and ``$S_{G}$ and $S_{D}$'' are calculated as follows:
\begin{align}
T_{-} &= \frac{1}{HW}\left|\sum_{i=1}^{H}\sum_{j=1}^{W}{T_{G}(i,j)}-\sum_{i=1}^{H}\sum_{j=1}^{W}{T_{D}(i,j)}\right|\\
S_{-} &= \frac{1}{HW}\left|\sum_{i=1}^{H}\sum_{j=1}^{W}{S_{G}(i,j)}-\sum_{i=1}^{H}\sum_{j=1}^{W}{S_{D}(i,j)}\right|
\label{eq:area_diff}
\end{align}
The balance between $T_{-}$ and $S_{-}$ is quantified by their harmonic mean as follows:
\begin{equation} 
{SEM}=\frac{2 T_{-} S_{-}}{T_{-}+S_{-}}
\label{eq:sae}
\end{equation}
The above SEM score gets smaller as the size-variable mask estimation works better.

\begin{figure*}[t]
  \begin{center}
    \includegraphics[width=\textwidth]{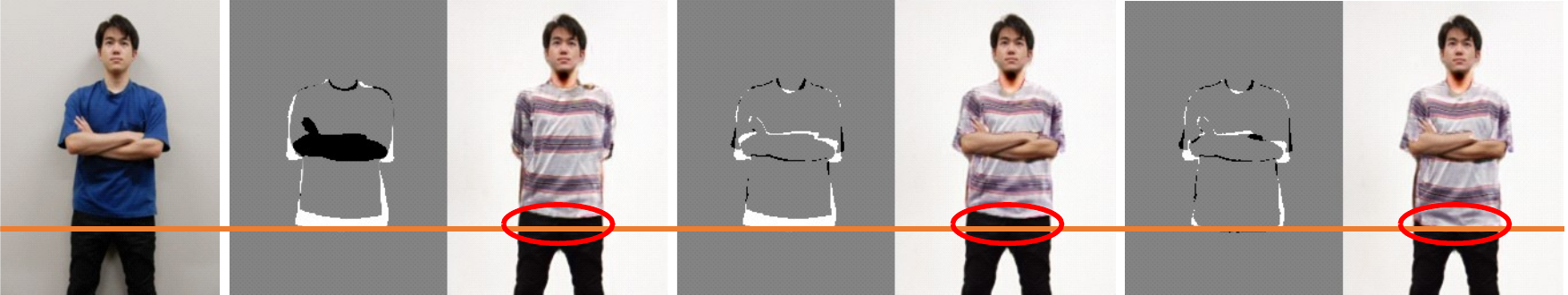}
  \end{center}
    \vspace{-4mm}
    \hspace*{5mm}
    $P_{R}$ (XL)~\hspace{12mm}
    $M_{G}$-$M_{D}$~\hspace{14mm}
    $P_{T}$~\hspace{15mm}
    $M_{G}$-$M_{D}$~\hspace{14mm}
    $P_{T}$~\hspace{15mm}
    $M_{G}$-$M_{D}$~\hspace{12mm}
    $P_{T}$\\
    ~\hspace*{40mm}
    (a) ACGPN~\cite{acgpn}~\hspace{25mm}
    (b) Ours($M_{R}$)~\hspace{30mm}
    (c) Ours
  \caption{Visual comparison in comparative experiments. 
  In this example, M-size and XL-size clothes are used as clothes in the reference person and try-on clothes images.
  In (a) and (b), the torso hem and sleeves are not changed from $P_{R}$ to $P_{T}$. 
  In (c), on the other hand, the torso hem and sleeves in $P_{T}$ are extended in accordance with the physical size of the try-on clothes.
   }
  \label{fig:vis_comparison}
\end{figure*}


\section{Experiments}
\label{sec:experiment}


\subsection{Implementation Details}
\label{subsec:exp_imp}

We employ Graphonomy~\cite{graphonomy,human_parse} as SE. 
For estimating Densepose and key-points of each human body, G{\"{u}}ler~\textit{et al}.~\cite{densepose} and Cao~~\textit{et al}.~\cite{keypoint} are used in our experiments, respectively.
As for a warping-based virtual try-on method that accepts $M_{D}$, we used TPS and CFN in the pretrained ACGPN~\cite{acgpn}.
All these components for our network are modularized so that they can be replaced with the SOTA methods without difficulty.
SFE consists of three full-connection layers, each of which has ReLe activation.
RMDN is an encoder-decoder network.
The encoder and decode consist of four and five convolutional layers, respectively.


\subsection{Evaluation Metrics and Dataset}
\label{subsec:exp_data_and_met}

\noindent{\bf Evaluation metrics.}
$M_{D}$ estimated by our size-variable mask deformation network is evaluated with SEM proposed in Sec.~\ref{subsec:prop_eval}.
Furthermore, the try-on image $P_{T}$ is also evaluated with Learned Perceptual Image Patch Similarity (LPIPS)~\cite{lpips} and Frechet Inception Distance (FID)~\cite{fid}, both of which are widely used in the field of image generation to evaluate the perceptual image quality.

\begin{table}[t]
  \caption{Comparison of BaseDataset and ProjDataset.}
  \label{table:comp_datasets}
  \centering
  \begin{tabular}{l||cccc} \hline
    Method & SEM($\times10^2$)↓ & LPIPS & FID \\\hline \hline
    BaseDataset & 0.49  & \red{0.44} & 16.83 \\\hline\hline
    ProjDataset (Ours) & \red{0.42} & \red{0.44} & \red{16.25} \\\hline
    \end{tabular}
\end{table}

\noindent{\bf Dataset.}
Our size-variable virtual try-on dataset, which is proposed in Sec.~\ref{subsec:prop_dataset}, is used.
For training in all experiments, person images $P_{R}$ and $P_{G}$ are augmented by flip and rotation.
The two types of datasets (i.e., BaseDataset and ProjDataset) are compared to validate the effectiveness of our proposed dataset generation method.
For this validation, the performance of virtual try-on is considered to be the measure of the dataset quality.
That is, $M_{D}$ and $P_{T}$ as the outputs of our method are evaluated by ``SEM'' and ``LPIPS and FID,'' respectively.

The results are shown in Table.~\ref{table:comp_datasets}. 
Since we can see that ProjDataset is better, ProjDataset is used in all experiments in what follows.

\begin{table}[t]
  \caption{Quantitative comparison. 
  The best and second-best results in each column are colored in \red{red} and \blue{{\bf blue}}. While StyleGAN is an IST-based method, other methods are warping-based methods.
  }
  \label{table:result_final}
  \centering
  \begin{tabular}{c||cccc} \hline
     Method & SEM($\times10^2$) $\downarrow$ & LPIPS $\downarrow$ & FID $\downarrow$ \\\hline \hline
     StyleGAN~\cite{stylegan} & 1.40 & 0.43 & 84.66 \\ \hline\hline
     ACGPN~\cite{acgpn} & 1.03 & 0.45 & 35.10 \\\hline
     Ours($M_{R}$) & \blue{{\bf 0.73}} & \red{0.44} & \red{15.77} \\\hline\hline
     Ours & \red{0.42}  & \blue{{\bf 0.44}} & \blue{{\bf 16.25}} \\\hline
    \end{tabular}
\end{table}

\begin{figure*}[t]
  \begin{minipage}{0.05\textwidth}
    Ours \\w/o \\RMDN\\~\vspace{15mm}\\
    Ours \\w/o \\MR\\~\vspace{15mm}\\
    Ours\\
  \end{minipage}
  \begin{minipage}{0.93\textwidth}
    \includegraphics[width=\textwidth]{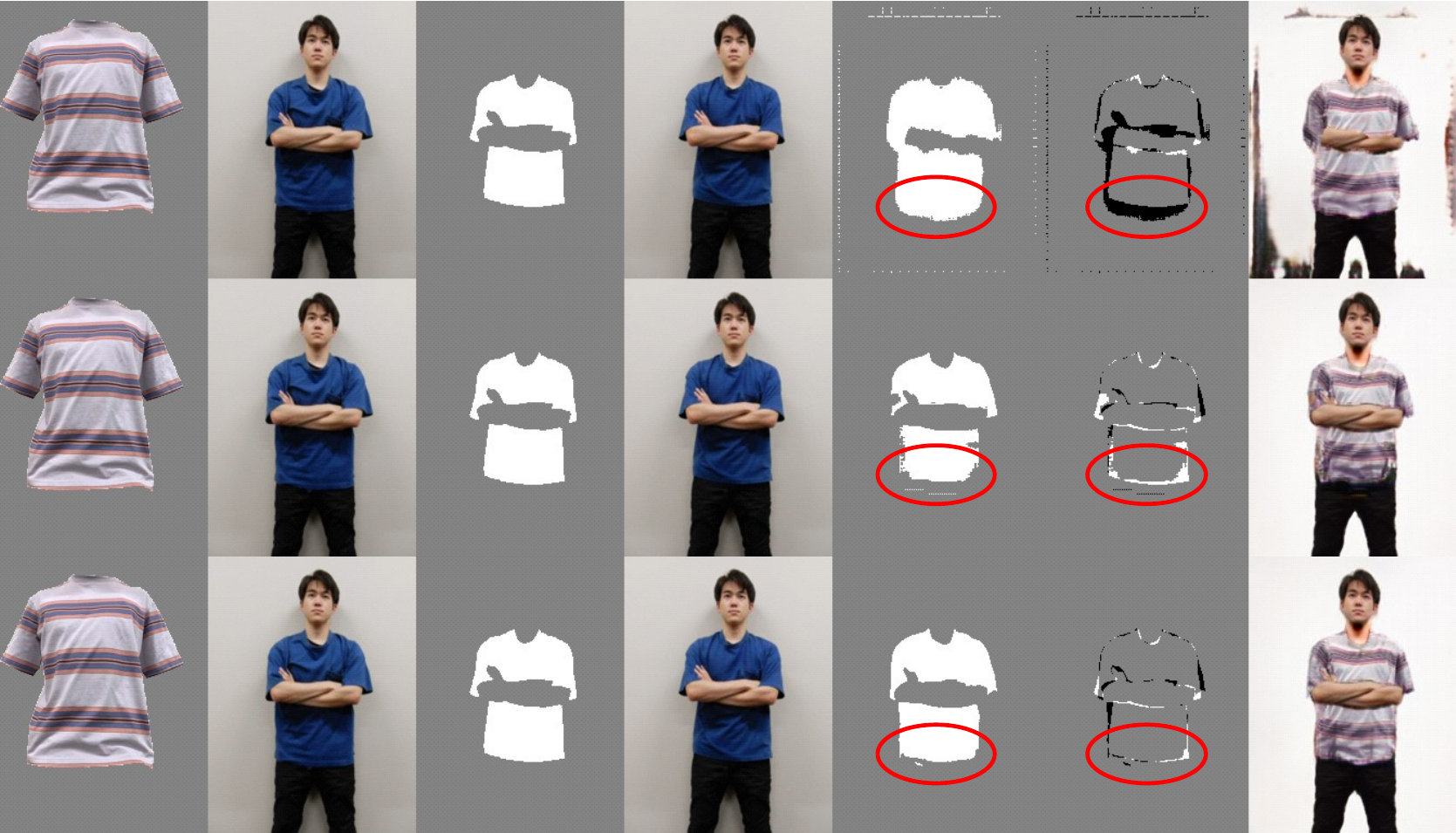}
  \end{minipage}
  \hspace*{22mm}
  $C_{T}$\hspace{19mm}
  $P_{R}$\hspace{16mm}
  $M_{G}$\hspace{16mm}
  $P_{G}$\hspace{16mm}
  $M_{D}$\hspace{12mm}
  $M_{D}$ - $M_{G}$\hspace{14mm}
  $P_{T}$
  \caption{Visual results for validating the effectiveness of RMDN and MR in ablation studies.
  }
  \label{fig:vis_ablation}
\end{figure*}


\subsection{Comparative Experiments}
\label{subsec:exp_comparative}

Note that, since all existing virtual try-on methods have no function for changing the cloth size, it is impossible to show fair comparative experiments in terms of the performance on the virtual try-on task.
However, to validate the necessity of the function for changing the cloth size, our method is compared with ACGPN~\cite{acgpn} as a general virtual try-on method that cannot change the cloth size.
ACGPN is selected because (i) it is one of the SoTA methods for warping-based virtual try-on and (ii) since ACGPN and our method have the same networks of TPS and CFN, the difference between these two methods is how to produce the mask $M_{D}$.
Therefore, a comparison between ACGPN and our method clearly validates the effectiveness of our proposed size-variable MDN.
In addition to ACGPN, the effectiveness of our size-variable MDN is verified by using $M_{R}$ instead of $M_{D}$ deformed by the size-variable MDN as the input for TPS and CFN.
This method is called Ours($M_{R}$).
While all the above methods are warping-based try-on methods, StyleGAN~\cite{stylegan} as an IST-based method is also evaluated so that style parameters were manually optimized for changing the clothes size.

Quantitative results are shown in Table~\ref{table:result_final}. 
Our method outperforms the other methods on SEM. 
The SEM scores of the previous methods~\cite{stylegan,acgpn} are inferior to Ours because the physical clothes size is not directly given to those methods.
This result demonstrates that our size-variable MDN can deform the mask according to the physical size of the clothes.
This is the biggest goal of this work.

The visual quality of the generated virtual try-on images is also important.
Ours and Ours($M_{R}$) are the best on LPIPS, and Ours is the second-best on FID, while the gap between Ours($M_{R}$) and Ours is not large (i.e., 15.77 vs. 16.25).
The superiority of Ours($M_{R}$) may be because the reality of $M_{D}$ generated by our size-variable MDN cannot reach that of the real clothes silhouette (i.e., $M_{R}$) even if our size-variable MDN is optimized by the adversarial loss.
FID of StyleGAN is much inferior to the other methods, probably because of the difficulty in hand-tuning of style parameters for changing the clothes size

Visual results are shown in Fig.~\ref{fig:vis_comparison}. 
The orange auxiliary line is located along the torso hem of the clothes in $P_{R}$.
Since the clothes sizes of the reference person image $P_{R}$ and the try-on clothes are M-size and XL-size, respectively, the torso hem and sleeves should be extended in the try-on image $P_{T}$.
With (a) ACGPN~\cite{acgpn} and (b) Ours($M_{R}$), it can be seen that the torso hem and sleeves are not changed.
(c) Our method, on the other hand, can extend the torso hem and sleeve in $P_{T}$.
This result clearly demonstrates that our method can achieve size-variable virtual try-on in $P_{T}$ different from the other methods.

\begin{table}[t]
  \caption{Ablation studies of our method. 
  }
  \label{table:ablation}
  \centering
  \begin{tabular}{l||ccc} \hline
    Method & SEM($\times10^2$) $\downarrow$ & LPIPS $\downarrow$ & FID $\downarrow$ \\\hline \hline
    Ours w/o RMDN & 1.21 & 0.54 & 67.51 \\\hline
    Ours w/o MR & 0.78 & \red{0.44} & 19.92 \\\hline
    Ours w/o $P_{R}$ & \blue{{\bf 0.44}} & \red{0.44} & \red{15.94} \\\hline\hline
    Ours  & \red{0.42} & \red{0.44} & \blue{{\bf 16.25}}\\\hline
    \end{tabular}
\end{table}


\subsection{Ablation Study}
\label{subsec:exp_eblation}

The effectiveness of each component in our method is verified by ablating the following three components from our proposed method.
We ablate the Residual Mask Deformation Network (RMDN) by directly estimating $M_{D}$ from $P_{R}$ and $M_{R}$ without the residual connection.
%
We also ablate the Mask Refiner (MR) in which $RM_{D}$ is added with $M_{R}$ to estimate $M_{D}$ without the MR.
Furthermore, $P_{R}$ given to RMDN is also ablated.

The quantitative results are shown in Table \ref{table:ablation}.
Our method is the best on SEM and LPIPS.
The improvements on SEM validate that RMDN, MR, and $P_{R}$ certainly improve the virtual try-on quality in terms of the clothes size.
In particular, the large gap on SEM between ``Ours'' and ``Ours w/o RMDN'' reveals that RMDN successfully estimates $M_{D}$ from the clothes sizes.
The difference between ``Ours'' and ``Ours w/o MR'' shows that the mask refiner allows fine adjustment of $M_{D}$, especially in the boundaries, rather than just elementwise adding $RM_{D}$ to $M_{R}$.

As for FID, the difference between ``Ours'' and ``Ours w/o $P_{R}$'' is not significant, while Ours is the second-best.
This result is also demonstrated in visual results shown in Fig.~\ref{fig:vis_ablation}.
The first, second, and third rows show the results obtained by ``Ours w/o RMDN,'' ``Ours w/o MR,'' and ``Ours,'' respectively.
Compared with the other methods, Ours can spatially align $M_{D}$ with $P_{R}$, as shown in the regions enclosed by the red ellipses in $M_{D}$ and $M_{D}-M_{G}$; residual pixels in $M_{D} - M_{G}$ are decreased as $M_{D}$ becomes better.
In ``Ours w/o RMDN,'' noisy lines appeared in $M_{D}$ give a negative impact on $P_{T}$. 
Such noisy $M_{D}$ degrades the perceptual quality of $P_{T}$, as also shown in Table~\ref{table:ablation}.

\subsection{Detailed Analysis}
\label{subsec:exp_det}


\begin{figure}[t]
  \begin{minipage}{0.05\columnwidth}
  {\scriptsize
    \vspace{8mm}
    $\mathcal{L}_{B}$\\~\vspace{7mm}\\
    $\mathcal{L}_{D}$\\~\vspace{4mm}\\
    $\mathcal{L}_{W}$\\ (Ours)
  }
  \end{minipage}
  \begin{minipage}{0.92\columnwidth}
    \includegraphics[width=\columnwidth]{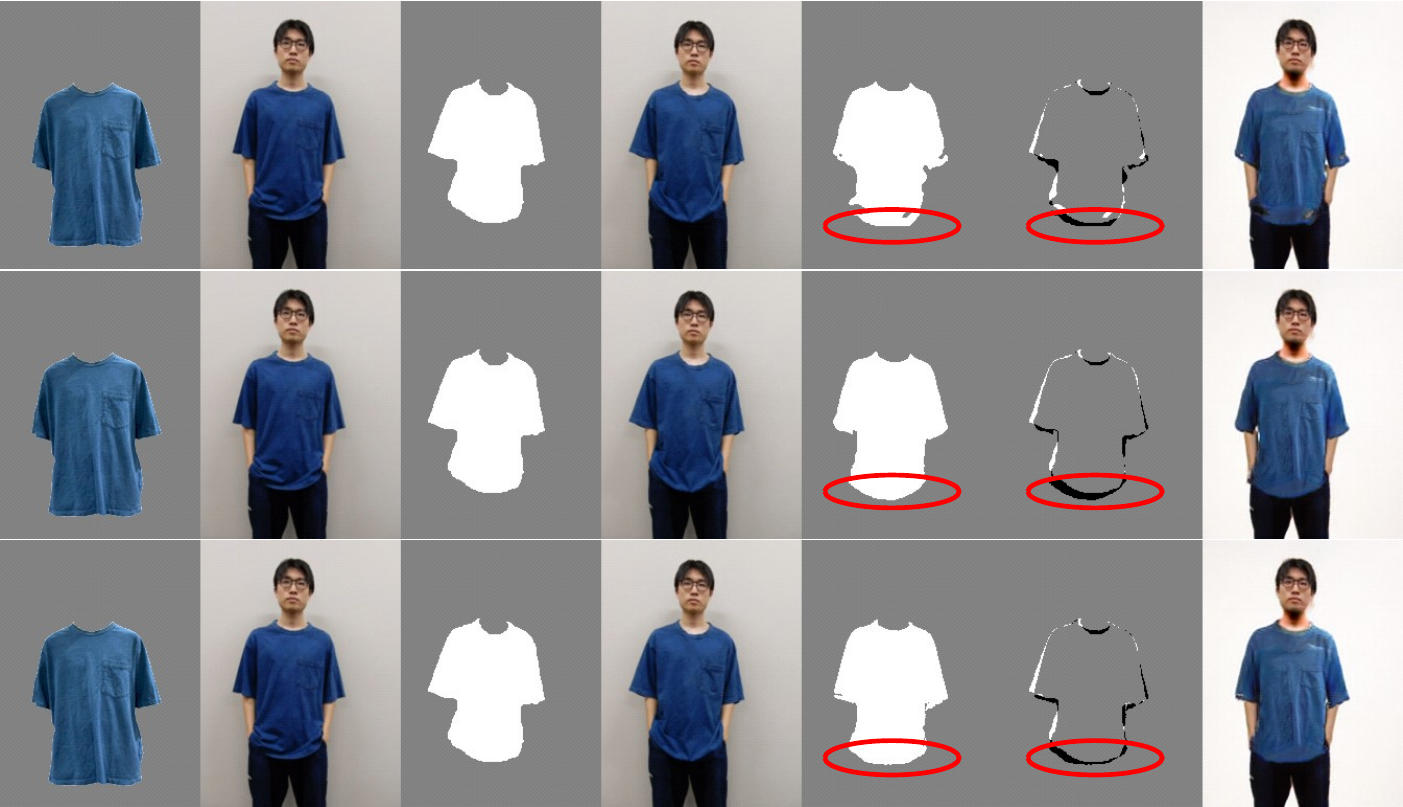}
  \end{minipage}
  \hspace*{12mm}
  {\footnotesize
  $C_{T}$
  \hspace{5mm}
  $P_{R}$
  \hspace{5mm}
  $M_{G}$
  \hspace{4mm}
  $P_{G}$
  \hspace{4mm}
  $M_{D}$
  }
  \hspace{1mm}
  {\scriptsize
  $M_{D}$ - $M_{G}$
  }
  \hspace{1mm}
  {\footnotesize
  $P_{T}$
  }
  \caption{Visual comparison of loss functions for comparing $M_{D}$ with $M_{G}$.
  }
  \label{fig:vis_det_mask_loss}
\end{figure}

\begin{table}[t]
  \caption{
  Comparison of loss functions for comparing $M_{D}$ with $M_{G}$.
  }
  \label{table:comp_loss_func_mask}
  \centering
  \begin{tabular}{l||ccc} \hline
    Loss function & SEM($\times10^2$) $\downarrow$ & LPIPS $\downarrow$ & FID $\downarrow$ \\\hline \hline
    $\mathcal{L}_{B}$ & 0.46 & \red{0.44} & \blue{{\bf 16.48}} \\\hline
    $\mathcal{L}_{D}$ & \blue{{\bf 0.45}} & \red{0.44} & 17.27 \\\hline\hline
    $\mathcal{L}_{W}$ (Ours) & \red{0.42} & \red{0.44} & \red{16.25}\\\hline
    \end{tabular}
\end{table}

\noindent{\bf Comparison of loss functions for comparing $M_{D}$ with $M_{G}$:}
While the Weighted Binary Cross Entropy loss, $\mathcal{L}_{W}$, is employed for comparing $M_{D}$ with $M_{G}$
as expressed in (\ref{eq:loss}) in our method, $\mathcal{L}_{W}$ is replaced by the Binary Cross Entropy loss, $\mathcal{L}_{B}$, or the Dice loss, $\mathcal{L}_{D}$ for comparative experiments.
The quantitative results are shown in Table~\ref{table:comp_loss_func_mask}.
Our method with $\mathcal{L}_{W}$ achieves the best performance in all metrics. 
This result shows that $\mathcal{L}_{W}$ can properly weight zero and non-zero pixels in $M_{D}$ and $M_{G}$ for adjusting the clothes size while maintaining the visual reality of the clothes silhouette.
The visual comparison is shown in Fig.~\ref{fig:vis_det_mask_loss}.
Compared with the other methods, $\mathcal{L}_{W}$ (Ours) successfully estimates $M_{D}$, as shown in the regions enclosed by the red ellipses in $M_{D}$ and $M_{D}-M_{G}$.


\begin{figure}[t]
  \begin{minipage}{0.08\columnwidth}
  {\scriptsize
    \vspace{8mm}
    $\mathcal{L}_{MAE}$\\
    ~\vspace{7mm}\\
    $\mathcal{L}_{MSE}$\\
    ~\vspace{4mm}\\
    $\mathcal{L}_{D}$\\ (Ours)
  }
  \end{minipage}
  \begin{minipage}{0.90\columnwidth}
    \includegraphics[width=\columnwidth]{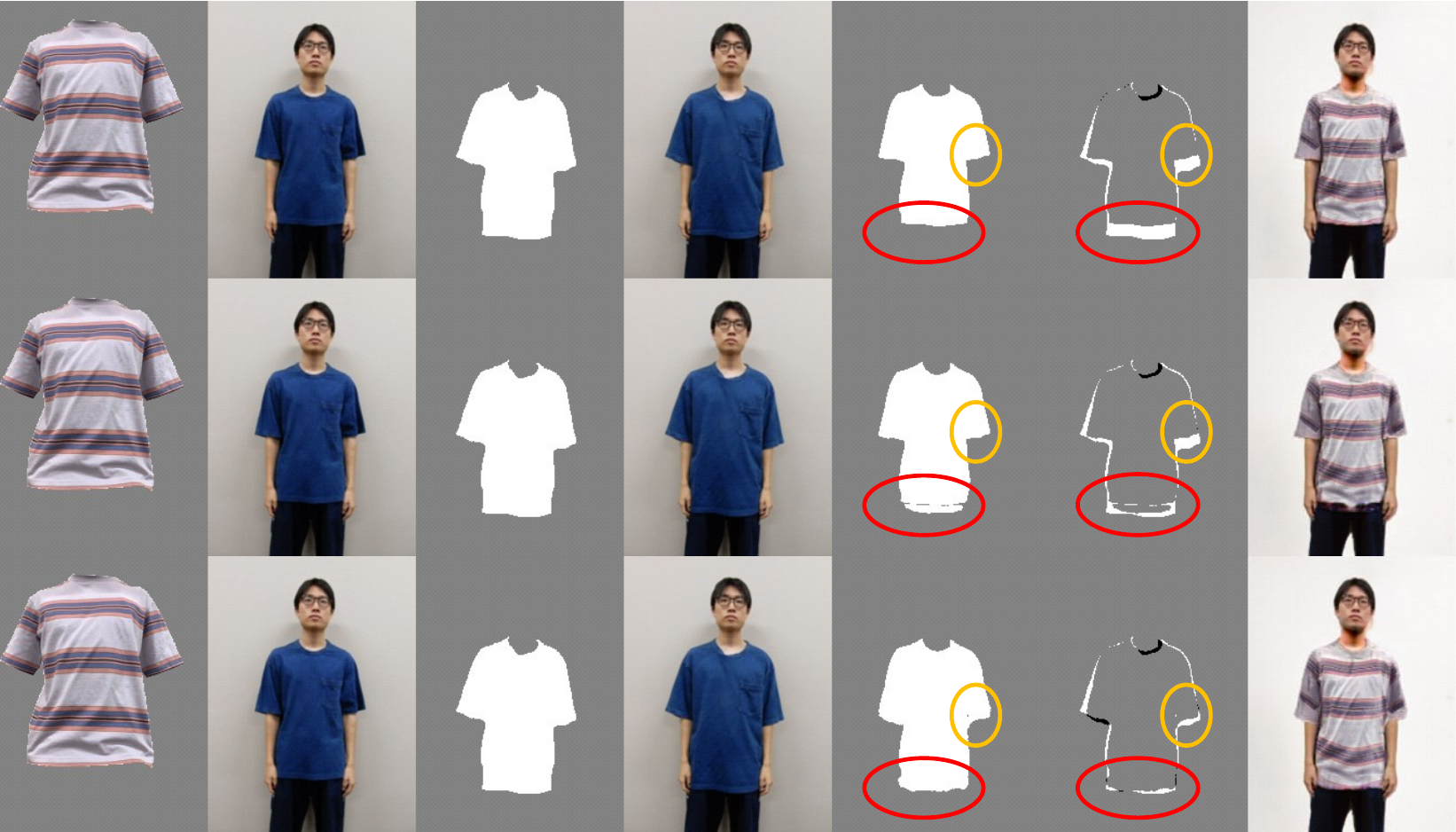}
  \end{minipage}
  \hspace*{12mm}
  {\footnotesize
  $C_{T}$
  \hspace{5mm}
  $P_{R}$
  \hspace{5mm}
  $M_{G}$
  \hspace{4mm}
  $P_{G}$
  \hspace{4mm}
  $M_{D}$
  }
  \hspace{1mm}
  {\scriptsize
  $M_{D}$ - $M_{G}$
  }
  \hspace{1mm}
  {\footnotesize
  $P_{T}$
  }
  \caption{Visual comparison of loss functions for comparing $RM_{D}$ with $RM_{G}$.
  }
  \label{fig:vis_det_res_loss}
\end{figure}

\begin{table}[t]
  \caption{Comparison of loss functions for $RM_{D}$ and $RM_{G}$.
  While $\mathcal{L}_{Dice}$ is used for $RM_{D}$ and $RM_{G}$ in our method, $\mathcal{L}_{MAE}$ and $\mathcal{L}_{MSE}$ are compared in this experiment.
  }
  \label{table:comp_loss_func_res_mask}
  \centering
  \begin{tabular}{l||ccc} \hline
    Loss function & SEM($\times10^2$) $\downarrow$ & LPIPS $\downarrow$ & FID $\downarrow$ \\\hline \hline
    $\mathcal{L}_{MAE}$ & 0.72 & \red{0.44} & \red{15.30} \\\hline
    $\mathcal{L}_{MSE}$ & \blue{{\bf 0.51}} & \red{0.44} & \blue{{\bf 15.82}} \\\hline\hline
    $\mathcal{L}_{Dice}$ (Ours) & \red{0.42} & \red{0.44} & 16.25\\\hline
    \end{tabular}
\end{table}

\noindent{\bf Comparison of loss functions for comparing $RM_{D}$ with $RM_{G}$:}
The loss function for comparing $RM_{D}$ with $RM_{G}$, the Dice loss $\mathcal{L}_{D}$ in our method is replaced by alternative loss functions. 
$\mathcal{L}_{D}$ is replaced by the Mean Absolute Error $\mathcal{L}_{MAE}$ or the Mean Squared Error $\mathcal{L}_{MSE}$.
As shown in Table~\ref{table:comp_loss_func_res_mask}, 
our method outperforms the other methods on SEM, as with all the other experiments shown before.
The big improvements in the clothes size can also be validated in the visual results, as shown in the red and orange ellipses in Fig.~\ref{fig:vis_det_res_loss}.
Regarding LPIPS and FID, the LPIPS scores of all the methods are equal, while ours is inferior to the others.
However, it is difficult to see any remarkable difference in the visual results, as shown in Fig.~\ref{fig:vis_det_res_loss}.


\section{Conclusion}
\label{sec:conclusion}

This paper proposed the size-variable mask deformation network, the size-variable virtual try-on dataset, and the size evaluation metric for size-variable virtual try-on, which is a new problem in this research area.
Our proposed mask deformation network can estimate the mask in accordance with the physical size of the try-on clothes. The results of size-variable virtual try-on are evaluated by our size evaluation metric in which the lengths of the torso hem and sleeves are particularly evaluated.
Experimental results demonstrate that our method outperforms the other methods quantitatively regarding the performance of size-variable virtual try-on; our method is the best on the size evaluation metric (SEM), with a large margin improvement in all the experiments.
In the visual quality evaluated by LPIPS and FID also, our method is comparable with the other methods.
Furthermore, various visual results also validate the effectiveness of our proposed method.

%
Extending our dataset is important future work because its scale is not sufficient yet to validate the performance with more subjects, more poses, and more clothes, while our dataset is the first one that can be used for the size-variable virtual try-on task.
While a general pose estimation method~\cite{keypoint} is used in our experiments, for key-point estimation under clothes, human pose estimation should be optimized for this purpose~\cite{DBLP:conf/eccv/UkitaTK08,DBLP:conf/iccv/UkitaHK09}.
Active learning also benefits efficient and accurate pose estimation in a query video~\cite{DBLP:conf/wacv/TaketsuguU24}.
Another future work is to verify the modularity of our method by replacing a component/sub-network.
For example, it is not guaranteed that existing TPS can always generate authentic results for try-on clothes. 
Since our size-variable mask deformation network is modular, TPS can be replaced by any SoTA texture rendering method.

{\small
\bibliographystyle{ieee_fullname}
\bibliography{egbib}
}

\end{document}